\documentclass{article} % For LaTeX2e
\usepackage{iclr2026_conference,times}

\usepackage[T1]{fontenc}  % correct rendering of ž in author name
\usepackage{booktabs}     % \toprule, \midrule, \bottomrule
\usepackage{graphicx}     % \includegraphics
\usepackage{amsmath}      % \text inside math, equation environment
\usepackage{xcolor}
\usepackage{enumitem}     % list spacing configuration below
\usepackage{microtype}    % protrusion and font expansion
\usepackage{url}
\usepackage{hyperref}
\usepackage[capitalize,noabbrev]{cleveref}  % must be loaded after hyperref

\setlist[itemize]{
  topsep      = 0.5\baselineskip,
  itemsep     = 0.5\baselineskip,
  parsep      = 0pt,
  partopsep   = 0pt,
  leftmargin  = 1.4em,
  labelsep    = 0.5em,
  label       = {\raisebox{0.15ex}{\scriptsize$\bullet$}}
}

\usepackage{amsmath,amsfonts,bm}

\def\eqref#1{equation~\ref{#1}}
\def\1{\bm{1}}

\DeclareMathAlphabet{\mathsfit}{\encodingdefault}{\sfdefault}{m}{sl}
\SetMathAlphabet{\mathsfit}{bold}{\encodingdefault}{\sfdefault}{bx}{n}

\title{Entangled Representations Amplify Collateral Damage in Unlearning}

\author{Evžen Wybitul \\
  University of Oxford \\
  \And
  Tim G. J. Rudner \\
  University of Toronto \\
  \And
  Christian Schroeder de Witt \\
  University of Oxford
}

\iclrfinalcopy % Uncomment for camera-ready version, but NOT for submission.
\begin{document}

\maketitle

\begin{abstract}
A long-held intuition in interpretability research is that representational entanglement---the sharing of structure between knowledge domains in a neural network---makes unlearning harder. While the intuition is widespread, it has never been directly tested in a controlled experiment. We present a way to do so: by repurposing Selective Gradient Masking (SGTM), we train a suite of six 254M-parameter language models on English Wikipedia with graded levels of disentanglement between biology and non-biology knowledge. Applying three standard unlearning methods to every model in the suite, we find that more disentangled models consistently achieve better retain--forget trade-offs: at a fixed level of forgetting, the most disentangled models incur roughly $4\times$ lower retain cost under two of the three methods, and $1.3\times$ lower under the third. Because our intervention changes only the model, not the data or the unlearning algorithm, this is direct evidence that representational entanglement is one of the causes of collateral damage in unlearning, as interpretability researchers have long suspected. A similar design could be used to test other structural claims from interpretability.
\end{abstract}

\begin{figure}[h]
  \centering
  \includegraphics[width=\linewidth]{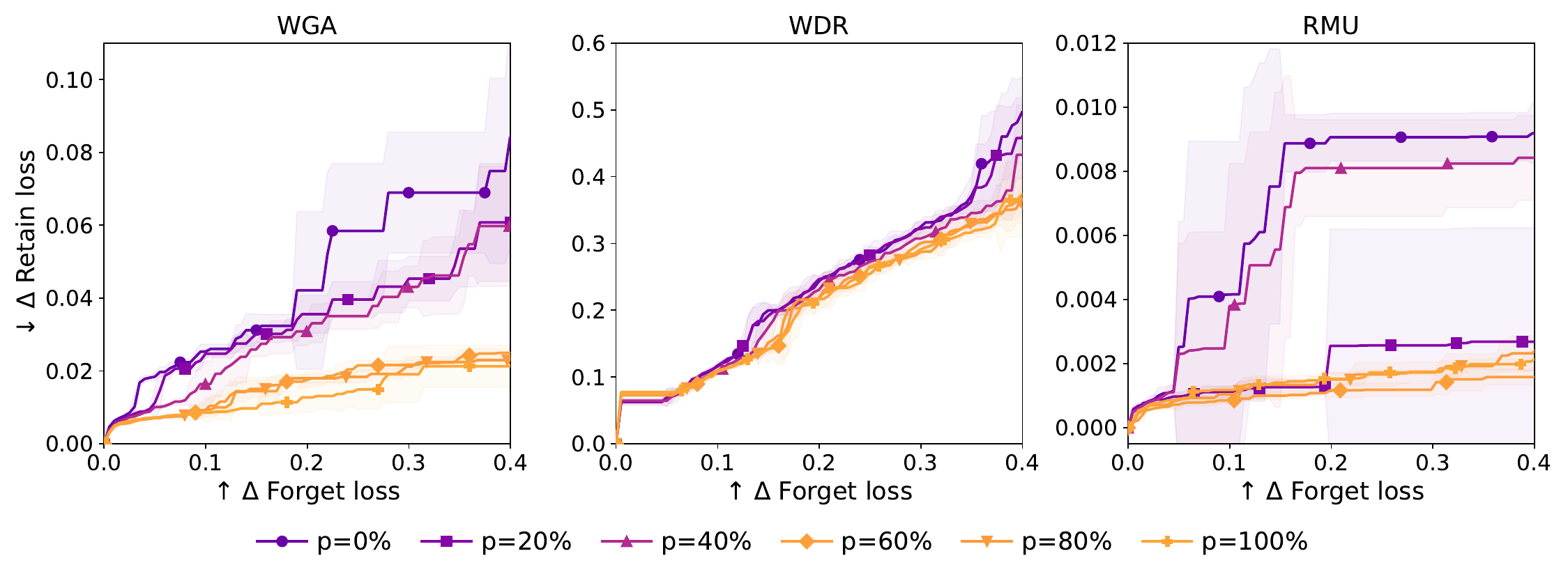}
  \caption{\textbf{The more disentangled the retain and forget domains are in a model, the better the retain--forget trade-off achieved by unlearning.}
    Retain--forget Pareto frontiers of three unlearning methods (WGA, WDR, RMU) across six models that differ in the fraction of SGTM training steps~($p$).
    Models are coloured by their Variance Entanglement Score (defined in~\cref{app:metrics}, shown in~\cref{fig:disentanglement}): purple means less disentangled, yellow more.
    Each frontier is constructed over four to six hyperparameter configurations; shaded regions show $\pm 1$ standard error across five seeds.
    Both axes are measured relative to each model's own pre-unlearning loss.
  Analysed in~\cref{sec:results}.}
  \label{fig:unlearning}
\end{figure}

\section{Introduction}

Broadly speaking, interpretability research is useful to the extent that it predicts what happens when we intervene on a network. Unlearning is a natural test case: interpretability has proposed two structural properties of neural networks that should, intuitively, affect how cleanly a targeted piece of knowledge can be unlearned. The first is \emph{localization}: where in the network's parameters the knowledge resides. The second is \emph{entanglement}: the degree to which the retain and forget domains share structure in the network---representations, processing pathways, or parameters. Both of these intuitions have taken a foothold in the unlearning literature~\citep{barez2025openproblemsmachineunlearning}, but only localization has so far been subjected to rigorous, controlled experimental testing~\citep{lee2025does, guo2025mechanistic, boglioni2026lacuna}. In this paper, we apply similar experimental methods to entanglement.

According to the entanglement intuition, high entanglement should make the retain--forget trade-off worse: if the knowledge to be forgotten and the knowledge to be retained share a lot of structure in the network, removing one without damaging the other should be harder. However, this has never been tested directly. Of the three main ingredients that go into unlearning---the model, the data, and the unlearning algorithm---current literature investigating entanglement has always held the model fixed, and tested the effect of entanglement only indirectly by varying the data composition~\citep{zhao2024what} or the unlearning algorithm~\citep{sondej2025collapseirrelevantrepresentationscir, tang2026logitslatentscontrastiverepresentation, chen2026clue}, leaving model-level entanglement confounded with other factors. The perhaps most natural way to test whether entanglement affects unlearning, namely, holding the datasets and algorithm fixed and varying only the degree of entanglement in the model, has so far not been attempted---in part because it requires a way to control entanglement rather than merely observe it.

Our main contribution is to fill this gap and do for entanglement what recent work has done for localization: build a controlled experiment that tests this intuition as directly as possible. We train a suite of six 254M-parameter language models on English Wikipedia~\citep{wikidump}, increasingly disentangling biology knowledge from the rest using Selective Gradient Masking (SGTM), an improved variant of gradient routing~\citep{shilov2025datafilteringknowledgelocalization, cloud2024gradientroutingmaskinggradients}, and verify with three metrics that the suite spans a graded range of disentanglement. On each model in the suite, we measure the retain--forget trade-off of three common unlearning methods: Weighted Gradient Ascent (WGA)~\citep{wang2025rethinking}, Weight Divergence Regularization (WDR)~\citep{siddiqui2025dormant}, and RMU~\citep{li2024wmdp}. \textbf{We find that more disentangled models consistently achieve better retain--forget trade-offs (\cref{fig:unlearning}):} at a fixed level of forgetting, the most disentangled models incur roughly $4\times$ lower retain cost than the most entangled one under WGA and RMU, and $1.3\times$ lower under WDR.

We want to be precise about what this establishes. Even in our experiments, we cannot change entanglement alone: we alter the training procedure, inducing changes to entanglement and, potentially, to other properties of the model. Nevertheless, the design rules out the data and algorithm confounds of earlier work. Taken together with the indirect evidence of earlier work, our results turn a long-standing intuition from interpretability into an empirical finding.

% ======================================================================
\section{Experimental Setup}

\subsection{Training Models with Varying Disentanglement}
\label{sec:training}

To train models in which the forget and retain domains are disentangled, we use a variant of gradient routing~\citep{cloud2024gradientroutingmaskinggradients} called Selective Gradient Masking~(SGTM)~\citep{shilov2025datafilteringknowledgelocalization}.
Although SGTM is a training method originally designed to make a group of parameters specialize on processing inputs from a target domain, we find we can repurpose it for training disentangled models.
Namely, when SGTM is used to specialize a group of parameters on the forget domain, the forget domain also grows more disentangled from retain, as we show in~\cref{sec:entanglement}.
We exploit this fact by varying the fraction of training steps in which we apply SGTM, producing models with different levels of disentanglement.

% --- Data ---

\paragraph{Data.}
Following the original SGTM paper, we train on English Wikipedia~\citep{wikidump} (approximately 3.7B~tokens), using article-level topic labels derived from Wikipedia's \texttt{articletopic} classifier~\citep{wikimedia_ores_articletopic}, and splitting the data into three domains: (1) \emph{forget}: all articles classified as STEM.Biology (${\sim}3.7\%$ of training tokens); (2) \emph{adjacent}: articles on topics closely related to biology (Medicine~\& Health, Chemistry, Earth~\& Environment); (3) \emph{retain}: all remaining articles, spanning Culture, Geography, History~\& Society, and the STEM topics unrelated to biology.

Each example is a 1,024-token block of article text.
We hold out separate test sets for each domain.
To reduce evaluation cost during our unlearning hyperparameter sweeps, we subsample $20\%$ of each test set and report all unlearning metrics on this fixed subsample.

% --- Model ---

\paragraph{Models.}
Following the original paper, all models we train are 254M-parameter GPT-Neo-style transformers with 16~layers, hidden dimension~1,024, 32~attention heads, and MLP dimension~4,096; the full configuration is given in~\cref{app:hyperparams}.
In each transformer block, we designate a small subset of parameters as the \emph{biology subnetwork}, $\theta_\text{bio}$: 1~attention head (out of~32) and 64~MLP hidden units (out of~4,096) per block.
We denote all remaining parameters by~$\theta_\text{other}$.

% --- SGTM ---

\paragraph{SGTM.}
As in the original paper, we assign training examples to one of three routing categories: (1) \emph{route-bio}: all forget (i.e., all biology) examples; (2) \emph{route-other}: a randomly sampled $10\%$ of retain and adjacent examples; (3) \emph{route-unchanged}: the remaining $90\%$ of retain and adjacent examples.

Depending on an example's routing category, SGTM modifies the training procedure during both the forward and backward passes.
For route-bio examples, gradients for~$\theta_\text{other}$ are zeroed after the backward pass, ensuring that biology knowledge flows only into~$\theta_\text{bio}$.%
\footnote{Unlike in the original paper, we do \emph{not} ablate~$\theta_\text{bio}$ after training.
We use SGTM solely for the purpose of controlling disentanglement, not for unlearning itself.}
For route-other examples, $\theta_\text{bio}$~is zeroed during the forward pass, training the model to perform well on non-target data even without the biology subnetwork.
Finally, route-unchanged examples update all parameters via standard training.
See~\cref{tab:interventions} for an overview.

\begin{table}[t]
  \centering
  \small
  \begin{tabular}{@{}lccc@{}}
    \toprule
    \textbf{Routing category} & \textbf{Forward pass} & \textbf{Backward pass} & \textbf{Updated params} \\
    \midrule
    route-bio & \textcolor{gray}{(standard)} & Zero $\nabla_{\theta_\text{other}}$ & $\theta_\text{bio}$ only \\
    route-other & Set $\theta_\text{bio} = 0$ & \textcolor{gray}{(standard)} & $\theta_\text{other}$ only$^{\dagger}$ \\
    route-unchanged & \textcolor{gray}{(standard)} & \textcolor{gray}{(standard)} & $\theta_\text{bio}$ and $\theta_\text{other}$ \\
    \bottomrule
  \end{tabular}
  \caption{Training interventions in SGTM, applied to each example based on its routing category.
  $^{\dagger}$\,$\theta_\text{bio}$ receives no gradient because its activations are zeroed during the forward pass.}
  \label{tab:interventions}
\end{table}

% --- Varying the fraction of SGTM steps ---

\paragraph{Varying the fraction of SGTM steps.}
In the original paper, SGTM was used in all training steps; we instead activate SGTM for only a fraction of the training steps, $p\%$.
We train six models with $p \in \{0, 20, 40, 60, 80, 100\}$;
the first $(100{-}p)\%$ of steps use standard training, and the remaining $p\%$ use SGTM as described in the previous paragraph.%
\footnote{Applying SGTM stochastically---routing each example with some probability rather than front-loading standard training and then switching to full SGTM---would produce substantial disentanglement even at low $p$, due to the \emph{absorption effect} described by~\citet{cloud2024gradientroutingmaskinggradients}.}

% ======================================================================
\subsection{Measuring Disentanglement between Domains}
\label{sec:measuring}

We measure disentanglement separately between all pairs of the three domains defined in~\cref{sec:training}: forget--retain, forget--adjacent, and, as a control, retain--adjacent.
Given a pair of domains, we sample 1,024 examples from their two test sets, then for each example we take the model's final-layer hidden states, average them over non-padding positions, and normalize the result to unit norm, yielding one vector per example.
We then compare the resulting two point clouds using three standard metrics: the Variance Entanglement Score (VES)~\citep{zhao2024what}, maximum mean discrepancy (MMD$^2$)~\citep{JMLR:v13:gretton12a}, and the sliced 2-Wasserstein distance (SW$_2^2$)~\citep{Bonneel2015}.
Lower VES and higher MMD$^2$ and SW$_2^2$ indicate more disentanglement; definitions and settings are given in~\cref{app:metrics}.

The three metrics are not on a common scale, and their absolute values are not comparable across domain pairs.
We therefore report, for each metric and each pair, the log ratio of the measured value to that of the $p{=}0\%$ model, so that every curve starts at zero and the quantity plotted is the relative change in separation induced by SGTM.

% ======================================================================
\subsection{Measuring the Retain--Forget Trade-off during Unlearning}
\label{sec:unlearning}

We apply three standard unlearning algorithms to each of the six models, using the forget and retain training data defined in~\cref{sec:training}.
Adjacent data is never used for optimization, only for evaluation.
We do not give the algorithms any information about how the models were trained.

Our interest is not in comparing methods against each other, but in checking whether, \emph{within each method}, more disentangled models yield better retain--forget trade-offs.
Since the six models differ in their losses before unlearning, we measure the \emph{change} in test loss relative to each model's own pre-unlearning value, $\Delta\ell_\text{forget}$ and $\Delta\ell_\text{retain}$.
For each method, we sweep over four to six hyperparameter configurations (see~\cref{app:unlearning-hyperparams} for the full list), and use the results to construct a retain--forget Pareto frontier for every model--method pair.
We run each configuration on five seeds for a method-specific number of optimizer steps, with early stopping when the forget loss exceeds that of a model trained with the biology data filtered out, plus a small buffer.%
\footnote{Beyond this point, further increases in forget loss reflect degradation of the model rather than removal of biology knowledge.}

\paragraph{Unlearning methods.}
Let $\operatorname{CE}(\theta;\, \mathcal{D})$ denote the mean per-token cross-entropy of model~$\theta$ on dataset~$\mathcal{D}$, and let $\theta_0$ denote the parameters before unlearning.
\begin{itemize}
  \item \textbf{Weighted Gradient Ascent (WGA)}~\citep{wang2025rethinking} combines gradient ascent on the forget set with gradient descent on the retain set, minimizing
    \begin{equation*}
      \mathcal{L}_\text{WGA}(\theta)
      = \operatorname{CE}(\theta;\, \text{retain})
      - \lambda_\text{WGA}\, \operatorname{CE}_\text{WGA}(\theta;\, \text{forget}),
      \quad \lambda_\text{WGA} > 0,
    \end{equation*}
    where $\operatorname{CE}_\text{WGA}$ is a reweighted cross-entropy that scales each token's log-probability loss by $[p_\theta(x_t \mid x_{<t})]^\alpha$, upweighting tokens the model predicts confidently so that gradient ascent targets the most well-learned predictions first.
    Each unlearning step draws one batch from the forget set and one from the retain set.

  \item \textbf{Weight Divergence Regularization (WDR)}~\citep{siddiqui2025dormant} fine-tunes on the retain set while maximizing the $L_2$ distance between the current parameters and $\theta_0$:
    \begin{equation*}
      \mathcal{L}_\text{WDR}(\theta)
      = \operatorname{CE}(\theta;\, \text{retain})
      - \lambda_\text{WDR}\, \sqrt{\frac{1}{|\theta|}\sum_i (\theta_i - \theta_{0,i})^2},
      \quad \lambda_\text{WDR} > 0.
    \end{equation*}

  \item \textbf{Representation Misdirection for Unlearning (RMU)}~\citep{li2024wmdp} modifies intermediate representations so that, on forget examples, activations at a designated layer are steered towards random target directions, while on retain examples, the same activations are anchored to their values under $\theta_0$.
    The loss is defined at a single layer, but the update is applied to the MLP weights of that layer and the two preceding it.
\end{itemize}

% ======================================================================
\section{Results}

\subsection{Models Trained with More SGTM Are More Disentangled}
\label{sec:entanglement}

\begin{figure}[t]
  \centering
  \includegraphics[width=\linewidth]{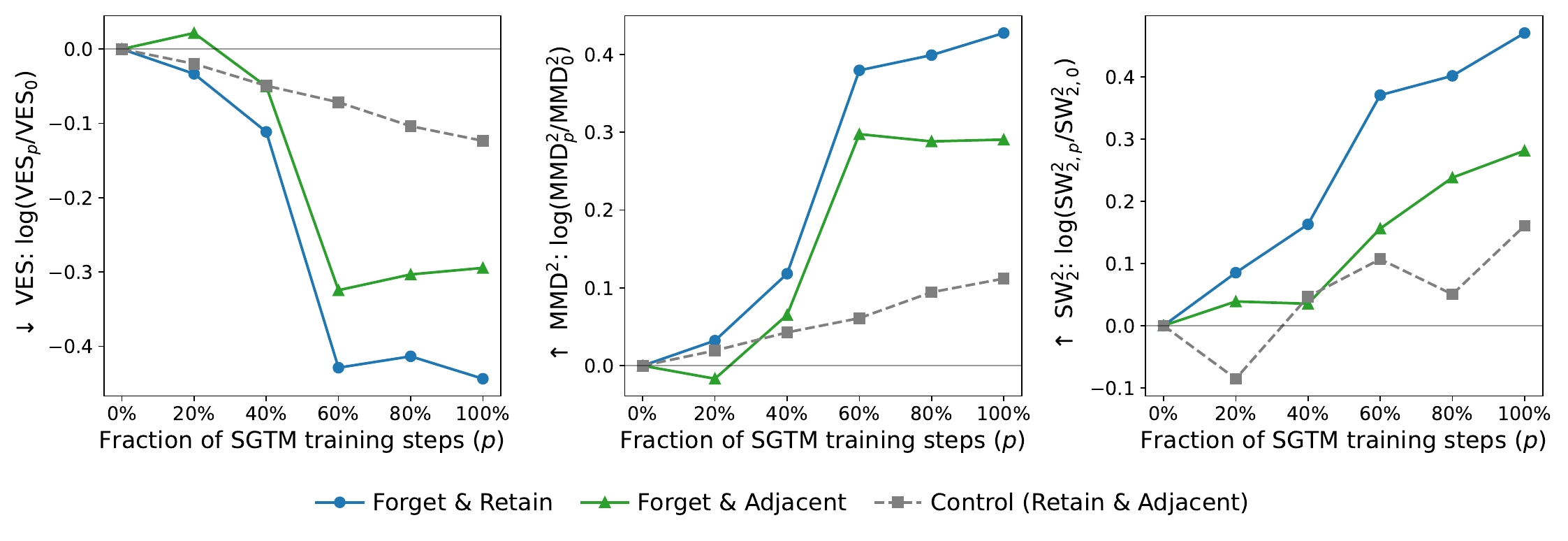}
  \caption{\textbf{The forget domain grows more disentangled from both retain and adjacent as the fraction of SGTM steps during training ($p$) increases}, with the largest change occurring between $p{=}40\%$ and $p{=}60\%$.
    The disentanglement of the retain--adjacent control also grows, but three to four times less.
    Each curve shows the log ratio of the measured value to that of the $p{=}0\%$ model, so all curves start at zero by construction, and the arrow on each axis gives the direction of increasing disentanglement.
  The three metrics are described in~\cref{app:metrics}.}
  \label{fig:disentanglement}
\end{figure}

We first verify that varying the fraction of SGTM training steps does produce models with meaningfully different levels of disentanglement.
All three metrics we use (see~\cref{sec:measuring}) agree that it does: as $p$ grows, the forget domain separates from both retain and adjacent (\cref{fig:disentanglement}).
The largest change happens between $p{=}40\%$ and $p{=}60\%$, and the models can be grouped into a less disentangled cluster ($p \leq 40\%$) and a more disentangled one ($p \geq 60\%$).

As a form of control, we also measure the disentanglement between retain and adjacent.
Recall that these two domains were grouped together during SGTM training (\cref{sec:training}), so we would expect their disentanglement to stay roughly constant across the model suite.
Instead, we observe that their disentanglement does grow with~$p$, but roughly three to four times less than in forget--retain.
We suspect this reflects the absorption effect~\citep{cloud2024gradientroutingmaskinggradients}: adjacent articles contain biology content that was never labelled as such, so that SGTM reshapes adjacent representations more than retain ones, and the two slightly drift apart.
Consistent with this, forget--adjacent disentangles about two-thirds as much as forget--retain.

\subsection{More Disentangled Models Achieve Better Retain--Forget Trade-offs}
\label{sec:results}

More disentangled models consistently achieve better retain--forget trade-offs, and this holds for all three unlearning methods (\cref{fig:unlearning}).
The size of the effect differs substantially between methods, but its direction does not.

\paragraph{Retain--forget results.}
All comparisons below are made at $\Delta\ell_\text{forget} = 0.4$.
\begin{itemize}
  \item \textbf{WGA:} The effect is clearest here.
    The most disentangled models ($p \geq 60\%$) incur roughly $4\times$ lower retain cost than the most entangled ($p{=}0\%$) model, which has the highest retain cost at every level of forgetting.
    The frontiers also separate into the same two clusters we found in~\cref{sec:entanglement}: the less disentangled models ($p \leq 40\%$) form one, and the more disentangled ones ($p \geq 60\%$) the other.

  \item \textbf{WDR:} The trend is in the same direction but weaker.
    The most disentangled models incur roughly $1.3\times$ lower retain cost than $p{=}0\%$, but the frontiers sit close together and their standard errors overlap.

  \item \textbf{RMU:} The retain-loss scale is two orders of magnitude smaller than for WDR, reflecting RMU's minimal impact on retain performance overall.
    However, the effect persists even at this scale: the most disentangled models incur roughly $4\times$ lower retain cost than $p{=}0\%$.
    Again, the trend holds, except for $p{=}20\%$, which performs comparably to the disentangled cluster.
\end{itemize}

\paragraph{Adjacent--forget results.}
The adjacent--forget trade-off is the hardest case in our setup.
First, being the domain closest to forget, adjacent receives greater spillover damage from unlearning.
Second, the unlearning methods optimize on forget and retain data only, so adjacent knowledge is left undefended.
Even so, we observe the same ordering on adjacent as on retain for WGA and WDR; for RMU the results are less clear.
We report these results with additional discussion in~\cref{app:adjacent}.

% ======================================================================
\section{Conclusion}

Interpretability has lent two intuitions to the field of unlearning: that the difficulty of unlearning depends on where the forget knowledge is stored (localization) and on how much structure it shares with retained knowledge (entanglement). While the first has been rigorously tested in a controlled experimental setting, the second had not. In this paper, we fill this gap: holding the data and the unlearning algorithms fixed and varying only the model, we find that more disentangled models consistently achieve better retain--forget trade-offs across three unlearning algorithms. Analogously to the existing work on localization, our work helps turn another long-standing intuition from interpretability into an empirical finding, and the recipe---repurpose an off-the-shelf method to control a representational property during training, verify the control, and test its predicted consequence---could be applied to other structural claims from interpretability as well.

\bibliography{iclr2026_conference}
\bibliographystyle{iclr2026_conference}

% ======================================================================
\appendix
\crefalias{section}{appendix}
\crefalias{subsection}{appendix}
\crefname{appendix}{Appendix}{Appendices}
\Crefname{appendix}{Appendix}{Appendices}

\section{Training Hyperparameters}
\label{app:hyperparams}

See~\cref{tab:training-hyperparams} for the full training hyperparameters.

\begin{table}[htbp]
  \centering
  \small
  \begin{tabular}{@{}ll@{}}
    \toprule
    \textbf{Hyperparameter} & \textbf{Value} \\
    \midrule
    Parameters & 254M \\
    Layers & 16 \\
    Hidden dimension ($d$) & 1,024 \\
    Attention heads ($h$) & 32 \\
    MLP dimension ($d_\text{MLP}$) & 4,096 \\
    Context size & 1,024 \\
    Vocabulary size & 50,257 \\
    Tied embeddings & True \\
    Tokenizer & GPT-2 \\
    \midrule
    Optimizer & AdamW \\
    Learning rate & $6 \times 10^{-4}$ \\
    LR schedule & Cosine annealing with warmup \\
    Warmup steps & 1,000 \\
    Total training steps & 9,689 \\
    Batch size & 16 \\
    Weight decay & 0.1 \\
    $\beta_1$ & 0.9 \\
    $\beta_2$ & 0.95 \\
    \midrule
    Forget MLP dim ($d_\text{forget}$) & 64 \\
    Forget attention heads ($h_\text{forget}$) & 1 \\
    Confident retain fraction & 10\% \\
    Mask embeddings & True \\
    \bottomrule
  \end{tabular}
  \caption{Training hyperparameters, following~\citet{shilov2025datafilteringknowledgelocalization}.}
  \label{tab:training-hyperparams}
\end{table}

\section{Unlearning Hyperparameters}
\label{app:unlearning-hyperparams}

All unlearning methods use the AdamW optimizer with no weight decay.
Each configuration is run with five seeds (42--46).
Full configurations are given in \cref{tab:wga-hyperparams,tab:wdr-hyperparams,tab:rmu-hyperparams}.

\begin{table}[htbp]
  \centering
  \small
  \begin{tabular}{@{}ll@{}}
    \toprule
    \textbf{Hyperparameter} & \textbf{Value} \\
    \midrule
    \multicolumn{2}{@{}l}{\textit{Fixed}} \\
    Total steps & 50 \\
    Eval every & 5 steps \\
    $\alpha$ & 0.1 \\
    Retain loss coefficient & 1.0 \\
    \midrule
    \multicolumn{2}{@{}l}{\textit{Swept}} \\
    Learning rate & $\{1.25 \times 10^{-5},\; 2.5 \times 10^{-5},\; 5 \times 10^{-5}\}$ \\
    Forget loss coefficient ($\lambda_\text{WGA}$) & $\{0.128,\; 0.32\}$ \\
    \bottomrule
  \end{tabular}
  \caption{WGA hyperparameters (six configurations).}
  \label{tab:wga-hyperparams}
\end{table}

\begin{table}[htbp]
  \centering
  \small
  \begin{tabular}{@{}ll@{}}
    \toprule
    \textbf{Hyperparameter} & \textbf{Value} \\
    \midrule
    \multicolumn{2}{@{}l}{\textit{Fixed}} \\
    Total steps & 100 \\
    Eval every & 10 steps \\
    Retain loss coefficient & 1.0 \\
    \midrule
    \multicolumn{2}{@{}l}{\textit{Swept}} \\
    Learning rate & $\{10^{-4},\; 2 \times 10^{-4},\; 4 \times 10^{-4}\}$ \\
    Divergence coefficient ($\lambda_\text{WDR}$) & $\{3,\; 30\}$ \\
    \bottomrule
  \end{tabular}
  \caption{WDR hyperparameters (six configurations).}
  \label{tab:wdr-hyperparams}
\end{table}

\begin{table}[htbp]
  \centering
  \small
  \begin{tabular}{@{}ll@{}}
    \toprule
    \textbf{Hyperparameter} & \textbf{Value} \\
    \midrule
    \multicolumn{2}{@{}l}{\textit{Fixed}} \\
    Total steps & 100 \\
    Eval every & 5 steps \\
    Steering coefficient & 20 \\
    Retain anchoring coefficient & 100 \\
    Target layer & 7 \\
    Updated layers & 5, 6, 7 \\
    Updated parameters & MLP weights and biases \\
    Max gradient norm & 1.0 \\
    Early stopping min steps & 20 \\
    \midrule
    \multicolumn{2}{@{}l}{\textit{Swept}} \\
    Learning rate & $\{1.25 \times 10^{-5},\; 10^{-4},\; 8 \times 10^{-4},\; 6.4 \times 10^{-3}\}$ \\
    \bottomrule
  \end{tabular}
  \caption{RMU hyperparameters (four configurations).}
  \label{tab:rmu-hyperparams}
\end{table}

\section{Disentanglement Metrics}
\label{app:metrics}

We compare the two point clouds described in~\cref{sec:measuring} using the following three metrics.
\begin{itemize}
  \item \textbf{Variance Entanglement Score (VES)}~\citep{zhao2024what} compares within-group spread to between-group separation. Lower VES indicates more disentanglement.

  \item \textbf{Maximum Mean Discrepancy (MMD$^2$)}~\citep{JMLR:v13:gretton12a} is a standard nonparametric measure of the difference between two distributions.
    It maps both clouds into a reproducing kernel Hilbert space and measures the squared distance between their mean embeddings.
    We use the unbiased estimator and a sum of five Gaussian kernels with bandwidths $\sigma \in 0.82 \cdot \{0.5,\, 0.71,\, 1,\, 1.41,\, 2\}$.
    Higher MMD$^2$ indicates more disentanglement.

  \item \textbf{Sliced 2-Wasserstein (SW$_2^2$)}~\citep{Bonneel2015} is an optimal-transport distance between distributions, and the standard alternative to the 2-Wasserstein distance in high dimensions, where the latter is difficult to estimate reliably.
    It averages the squared 2-Wasserstein distance between projections of the two clouds onto random directions.
    We use 512 fixed random projections.
    Higher SW$_2^2$ indicates more disentanglement.
\end{itemize}

\section{Unlearning Trade-offs on Adjacent Knowledge}
\label{app:adjacent}

\begin{figure}[htbp]
  \centering
  \includegraphics[width=\linewidth]{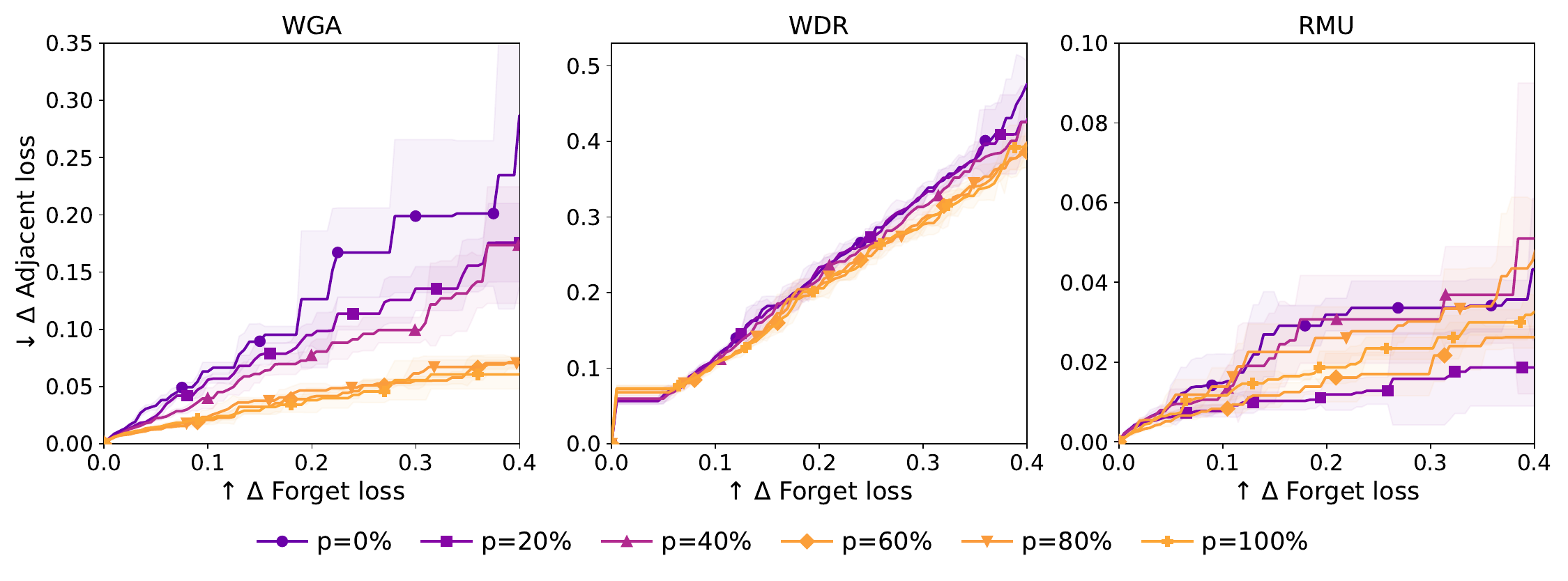}
  \caption{\textbf{Adjacent--forget Pareto frontiers, reproducing the ordering of~\cref{fig:unlearning} for WGA and WDR but not for RMU.}
    No unlearning method optimizes on adjacent data.
  Conventions are otherwise as in~\cref{fig:unlearning}.}
  \label{fig:unlearning-adjacent}
\end{figure}

\Cref{fig:unlearning-adjacent} shows the Pareto frontiers in the $(\Delta\ell_\text{adjacent},\, \Delta\ell_\text{forget})$ plane, constructed independently of the retain--forget frontiers of~\cref{fig:unlearning}.
As in~\cref{sec:results}, all comparisons are made at $\Delta\ell_\text{forget} = 0.4$.
\begin{itemize}
  \item \textbf{WGA:} The ordering from~\cref{sec:results} is reproduced, including the separation into two clusters.
    The most entangled model ($p{=}0\%$) incurs roughly $4\times$ the adjacent cost of the disentangled cluster ($p \geq 60\%$).

  \item \textbf{WDR:} The ordering is also reproduced, although, as in the retain--forget results, the frontiers are close together and their standard errors overlap.
    The means are ordered correctly across the suite.

  \item \textbf{RMU:} The frontiers largely overlap and we find no clear ordering.
    We do not have a conclusive explanation, but note that RMU's costs are by far the smallest of the three methods.
    We therefore take our results to mean that we cannot resolve an ordering at this scale of adjacent loss, rather than that there is none.
\end{itemize}

\end{document}